\documentclass{article}

\usepackage{arxiv}

\usepackage[utf8]{inputenc} 
\usepackage[T1]{fontenc}    
\usepackage{hyperref}       
\usepackage{url}            
\usepackage{booktabs}       
\usepackage{amsfonts}       
\usepackage{nicefrac}       
\usepackage{microtype}      
\usepackage{graphicx}
\usepackage{natbib}
\usepackage{doi}

\usepackage{subfigure}
\usepackage{multirow}
\usepackage{threeparttable}
\usepackage{subcaption} 
\usepackage{array} 
\usepackage{amsmath}

\usepackage{amsthm}
\newtheorem{theorem}{Theorem}

\usepackage{float} 
\usepackage[framemethod=TikZ]{mdframed} 
\mdfsetup{frametitlealignment=\center}
\makeatletter
\newenvironment{TextBox}[1]{
  \protected@edef\@currentlabelname{#1}
  \protected@edef\@currentlabel{#1}
  \begin{mdframed}[
    innerlinewidth=0.5pt,
    innerleftmargin=10pt,
    innerrightmargin=10pt,
    innertopmargin = 10pt,
    innerbottommargin=10pt,
    skipabove=\dimexpr\topsep+\ht\strutbox\relax,
    roundcorner=3pt,
    frametitle={#1},
    frametitlerule=true,
    frametitlerulewidth=1pt,
    backgroundcolor=gray!10]
}{
  \end{mdframed}
}
\makeatother
\usepackage{tikz}
\usepackage{makecell}
\usepackage[normalem]{ulem}
\useunder{\uline}{\ul}{}

\newcommand{\model}{GUNDAM}
\newcommand{\V}{$\mathcal{V}$ }
\newcommand{\E}{$\mathcal{E}$ }
\newcommand{\G}{$\mathcal{G}(\mathcal{V}, \mathcal{E})$}

\title{GUNDAM: Aligning Large Language Models with Graph Understanding}

\author{Sheng Ouyang \\
	Renmin University of China \\
	\texttt{ouyangsheng@ruc.edu.cn} \\
	\And
	Yulan Hu \\
	Renmin University of China \\
	\texttt{huyulan@ruc.edu.cn} \\
        \And
        Ge Chen \\
	University of Chinese Academy of Sciences \\
	\texttt{chenge221@mails.ucas.ac.cn} \\
        \And
	Yong Liu \\
	Renmin University of China \\
	\texttt{liuyonggsai@ruc.edu.cn} \\
}

\date{}

\renewcommand{\shorttitle}{GUNDAM}

\hypersetup{
pdftitle={GUNDAM: Aligning Large Language Models with Graph Understanding},
pdfauthor={Sheng Ouyang},
}

\begin{document}
\maketitle

\begin{abstract}
Large Language Models (LLMs) have achieved impressive results in processing text data, which has sparked interest in applying these models beyond textual data, such as graphs. In the field of graph learning, there is a growing interest in harnessing LLMs to comprehend and manipulate graph-structured data. Existing research predominantly focuses on graphs with rich textual features, such as knowledge graphs or text attribute graphs, leveraging LLMs' ability to process text but inadequately addressing graph structure. This work specifically aims to assess and enhance LLMs' abilities to comprehend and utilize the structural knowledge inherent in graph data itself, rather than focusing solely on graphs rich in textual content.
To achieve this, we introduce the \textbf{G}raph \textbf{U}nderstanding for \textbf{N}atural Language \textbf{D}riven \textbf{A}nalytical \textbf{M}odel (\model). This model adapts LLMs to better understand and engage with the structure of graph data, enabling them to perform complex reasoning tasks by leveraging the graph's structure itself. 
Our experimental evaluations on graph reasoning benchmarks not only substantiate that \model~ outperforms the SOTA baselines for comparisons. But also reveals key factors affecting the graph reasoning capabilities of LLMs. Moreover, we provide a theoretical analysis illustrating how reasoning paths can enhance LLMs' reasoning capabilities. 
\end{abstract}

\section{Introduction}
LLMs have achieved remarkable success in processing serialized natural language data~\cite{llama,gpt4,claude}. Recent explorations have pushed the boundaries of LLM applications beyond textual data, particularly into the realm of computer vision~\cite{visionllm,llava}. 
Parallel to these developments, the utilization of LLMs in graph representation learning has emerged as a focal point of interest~\cite{wang2023can,TAPE,OFA,Talk}. 


Considerable efforts have been devoted to deploying LLMs for processing graph data~\cite{huang2024can,llaga}. For instance, some work~\cite{OFA,TAPE} have utilized LLMs to handle rich-text graphs where node text attributes are used for node classification. Those work essentially translate the node classification task into a text classification task. While yielding impressive results, they primarily leverage the capability of LLMs to process textual features, consequently diminishing the emphasis on intrinsic graph data features, particularly the structural aspects of the graph. In light of this, we aim to explore a more fundamental and universal question: \textbf{Can LLMs genuinely understand graph data, especially the structural knowledge, and rely on it to perform complex reasoning tasks? }

Investigating whether LLMs can comprehend graph data is crucial, as graphs are fundamental data structures that represent entities and their complex interrelations, effectively modeling various real-world scenarios~\cite{wu2020graph,protein}. Enhancing LLMs' ability to understand graph data has the potential to significantly advance their general intelligence~\cite{sparks}. Many tasks, such as recommendation systems~\cite{recommendation} and knowledge graphs~\cite{relational}, explicitly or implicitly employ graph structures for reasoning processes. 
Recent advances~\cite{wang2023can,graphwiz,graphinstruct} have emphasized the exploration of LLMs' abilities in understanding and reasoning over graph structures. Specifically, benchmarks such as NLGraph~\cite{wang2023can} and GraphQA~\cite{Talk} have been developed to evaluate the performance of LLMs on various graph reasoning tasks. Studies utilizing closed-source models on these benchmarks indicate that while LLMs demonstrate fundamental competencies in addressing simple reasoning tasks, they struggle with more complex graph reasoning challenges, as observed by \citet{wang2023can} and \citet{Talk}.

Parallelly, the reasoning capabilities of open-source LLMs on graphs have garnered significant attention, with research primarily bifurcating into two categories. The first category focuses on leveraging prompt techniques to enhance the inherent reasoning abilities of LLMs for graph reasoning tasks. For instance, \citet{wang2023can} and \citet{Talk} explored advanced prompt strategies such as Chain of Thought (CoT) and Self-consistency (SC) to augment LLMs' reasoning efficacy. Additionally, \citet{let} introduced soft prompt techniques that map graph structures onto an LLM's embedding space for reasoning. These prompt-based approaches, though cost-effective due to negligible retraining or minimal parameter tuning, generally yield limited improvements in reasoning capabilities.
The second research avenue involves fine-tuning LLMs with synthetically generated graph reasoning data, which tends to significantly enhance their reasoning performance. However, this method has its drawbacks. First, the quality of synthetic data is unpredictable; manually constructing graph reasoning data requires considerable expertise in graph theory, making it a costly endeavor. Observations by \citet{wang2023can} and \citet{Talk} suggest that even powerful closed-source LLMs possess limited graph reasoning capabilities and thus cannot reliably generate high-quality graph reasoning data. Second, the specifics of what constitutes effective graph reasoning data and how to best utilize this data for maximizing an LLM’s capabilities in graph reasoning remain under exploration.

To this end, we introduce the \textbf{G}raph \textbf{U}nderstanding for \textbf{N}atural \textbf{L}anguage \textbf{D}riven \textbf{A}nalytical \textbf{M}odel (\model), which is designed to effectively comprehend graph structures and execute complex reasoning tasks over graphs. \model~integrates several pivotal components:
Firstly, to encode graph structures into the LLM, we employ the Graph Projection method. This approach effectively serializes the graph structure while preserving crucial information that is comprehensible to the LLM.
Secondly, we developed a pipeline to construct high-quality graph reasoning data, which includes CoT reasoning paths. This pipeline leverages graph algorithms not only to ensure accuracy but also to provide a detailed intermediate reasoning process. By systematically applying these algorithms, we can derive structured reasoning paths that enhance the depth and clarity of the model's reasoning ability.
Finally, we introduce an Alignment Tuning method, which fine-tunes \model~ using the graph reasoning data formulated. This tuning process aligns the model's reasoning capabilities with the structured graph reasoning data, further refining its ability to process and reason about graph-based information effectively.
The integration of these strategies enables \model~ to harness and expand upon the potential of LLMs in understanding and reasoning about complex graph structures.
The contributions of this paper are summarized as follows:
\begin{itemize}
    \item \textbf{Data Construction}: 
    We have developed a pipeline for constructing high-quality graph reasoning data that significantly enhances the reasoning capabilities of training data and improves model interpretability.

    \item \textbf{Methodology}: 
    We introduce \model, specifically designed to enable LLMs to comprehend graph-structured data and perform complex reasoning on graphs.

    \item \textbf{Understanding LLM Capabilities in Graph Reasoning}: Our experiments validate that \model~ achieves SOTA performance while also identifying factors that influence LLMs' graph reasoning capabilities. Additionally, we provide a theoretical analysis of how reasoning paths enhance LLMs' reasoning capabilities.
\end{itemize}

\section{Related Work}

\subsection{LLM for Graph}
In the realm of graph learning, substantial progress has been made by integrating Large Language Models (LLMs) with Text-Attributed Graphs (TAGs), where text descriptions are present either within the nodes or on the entire graph~\cite{OFA,ye2023natural,llaga,TAPE,labelfree}. In this context, several notable studies have emerged, focusing on diverse tasks such as node classification~\cite{TAPE,OFA,labelfree,llaga}, link prediction~\cite{lpnl}, and graph classification~\cite{zhao2023gimlet,qian2023can}. One pivotal area of application is node classification within citation networks, where TAPE~\cite{TAPE} stands out as a pioneering work. TAPE processes the titles and abstracts of research papers through an LLM to generate predictions and explanations. These outputs are subsequently utilized as augmented features for training a Graph Neural Network (GNN), allowing for improved classification performance. Addressing the challenge of sparse labeled data in graph datasets, \citet{labelfree} proposed an innovative pipeline that leverages LLMs to generate high-quality annotated data. This inventive approach reduces the annotation burden while enhancing the utility of available graph data. For handling graph data directly, LLaGA~\cite{llaga} proposes a novel approach by employing a soft prompt technique. It retains the pre-trained parameters of the LLM and introduces a Projector, which is trained to map node sequences into the token embedding space. This adjustment ensures that graph data are accommodated by the LLM for effective prediction tasks. For link prediction, LPNL~\cite{lpnl} introduces a scalable method leveraging LLMs. This method utilizes a two-stage sampling process for the source node and potential neighbor nodes to identify anchor nodes. Prompts generated based on these anchor nodes are then fed into the LLM for accurate prediction. In terms of classification, \citet{qian2023can} explore the use of LLMs for predicting molecular properties. This signifies an extension of LLM applicability to a broader range of graph types and data characteristics. OFA~\cite{OFA} represents a comprehensive approach that designs a versatile method applicable to various graph-related tasks. This method entails an intricate pipeline combining LLMs with GNNs. It makes use of subgraphs centered around nodes of interest (NOIs) and creates prompt graphs, which are then integrated into the original graph, thereby enriching the input for the GNN in an informed manner.



The application of LLMs to TAGs showcases both notable strengths and significant limitations. On one hand, LLMs exhibit significant potential in addressing complex textual data, effectively capturing intricate textual patterns and relationships, which are pivotal for tasks such as node classification. Their advanced capabilities in interpreting nuanced text significantly enhance classification accuracy. On the other hand, the integration of these models with graph data presents challenges. The conversion from graph structure to text sequences may result in verbose inputs that can be computationally intensive, thus hindering scalability. Moreover, there is a concern that the structural properties of graphs, such as topology and connectivity, are underutilized. This underutilization potentially limits the effectiveness of LLMs in fully leveraging the rich structural information inherent in graphs, which is crucial for a more comprehensive understanding and modeling of graph data.

\subsection{Reasoning on Graph}
Graph reasoning, an evolving field within graph representation learning and natural language processing, focuses on leveraging graph structures to perform cognitive tasks such as shortest path finding and topological sorting. Significant efforts~\cite{perozzi2024let,graphwiz,Talk,graphinstruct,graphllm} have been directed toward assessing and enhancing the reasoning capabilities of models over graph-structured data. The creation of benchmarks for graph reasoning has played a pivotal role in this research direction. For instance, NLGraph~\cite{wang2023can} and GraphQA~\cite{Talk} are notable projects that have independently established benchmarks tailored for evaluating reasoning tasks on graphs. These benchmarks primarily utilize closed-source models like GPT to explore the potential and limitations of LLMs in handling graph reasoning tasks. The initial explorations conducted by these benchmarks revealed that LLMs tend to underperform in fundamental graph reasoning tasks, underscoring a critical space for further development in model capabilities and training methods.

In an innovative approach to broaden the spectrum of graph reasoning challenges, VisionGraph~\cite{visiongraph} introduced a multimodal graph reasoning benchmark. This benchmark uniquely represents each graph as an image, thereby shifting the challenge towards visual Question Answering (QA) tasks. This approach not only diversifies the types of inputs that models must handle but also tests the adaptability of reasoning models in interpreting and processing information across differing data modalities. \citet{graphinstruct} introduced GraphInstruct, a benchmark specifically designed for instruction fine-tuning on graph reasoning tasks. This development paved the way for further innovations in the field, exemplified by GraphWiz~\cite{graphwiz}. GraphWiz represents the first open-source LLM dedicated to solving a variety of graph problems through explicit reasoning. In another notable development, \citet{let} proposed a method employing soft prompt technology that utilizes GNNs to map graph structures into the embedding space of LLMs. This technique allows LLMs to process graph data directly, thereby expanding their applicability in performing graph reasoning tasks. Additionally, significant work has been done in applying LLMs to TAGs for reasoning tasks. For instance, \citet{sun2023think} and \citet{rog} have focused on extracting genuine and effective relational paths within knowledge graphs. Their approaches aim to assist LLMs in generating faithful and interpretable reasoning outputs. By identifying and sourcing accurate relational paths, these models can generate more credible and logically consistent results in complex reasoning scenarios.

\section{Methodology}
\begin{figure*}[htb]
    \centering
    \includegraphics[width=0.8\linewidth]{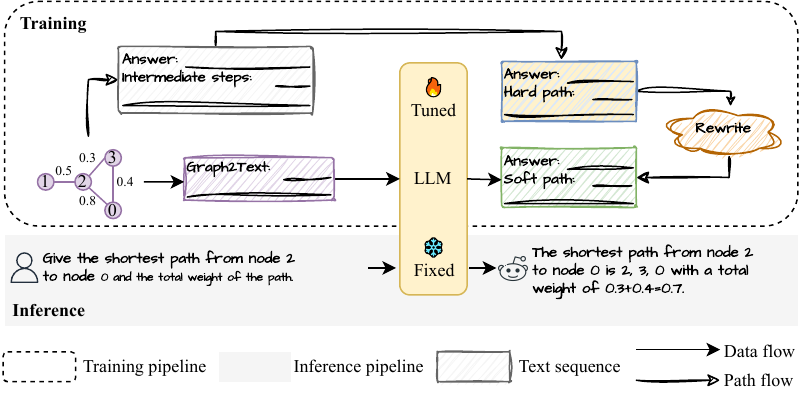}
    \caption{The overall framework of \model.}
    \label{fig:overview}
\end{figure*}

\subsection{Preliminaries} \label{sec:pre}
\paragraph{\textbf{Graph}}
A graph is denoted as \G, where \V and \E  represent the set of nodes and edges respectively. $(u, v, w) \in \mathcal{E}$ denotes an edge, where $u, v \in \mathcal{V}$ are the two nodes of the edge and $w$ denotes the weight of the edge. 

\paragraph{\textbf{Graph Reasoning}}
Graph reasoning is a fundamental component in the field of graph representation learning, where it focuses on inferring implicit relationships and attributes from nodes, edges, or subgraphs within structured graph data. This area involves critical operations such as information propagation, feature extraction, and pattern recognition, which are essential for supporting high-level reasoning and decision-making tasks across various applications including social network analysis~\cite{socialnet}, knowledge graph~\cite{sun2023think,rog}, etc.

In this paper, we specifically concentrate on graph reasoning that explicitly utilizes the structural aspects of graphs without relying on external textual data. Our focus is directed towards tasks such as identifying connectivity, performing topological sorting, finding shortest paths, and determining maximum flow within graphs. These tasks reveal the spatial and relational dynamics intrinsic to graph structures, thereby providing a deeper understanding of graph theory applications in practical scenarios. 



\subsection{Proposed Method}
In this section, we introduce the \model, illustrated in Figure ~\ref{fig:overview}, which outlines our strategy to empower an LLM to effectively process graph data inputs and undertake complex reasoning based on graph structures. To realize this goal, we need to address three pivotal challenges:
1) How can we encode graph structures for LLM input? We address this by adopting the Graph Projection method, which transforms graph structures into textual sequences that retain essential graph information in a format processable by the LLM.
2) How do we construct correct and diverse alignment data? We ensure data accuracy through the use of graph algorithms, while diversity is augmented by exploiting the generative capabilities of the LLM.
3) How can we enhance the graph reasoning capabilities of an LLM? This is achieved via Alignment Tuning, where we fine-tune the LLM specifically to enhance its performance on graph-based reasoning tasks.
Additionally, we theoretically investigate the benefits brought by CoT reasoning path. 

\paragraph{\textbf{Graph Projection}}
LLMs are not inherently equipped to process graph data directly. To facilitate LLMs in comprehending graph data and executing reasoning tasks, it is essential to first transform the graph data into a format that can be understood by LLMs. Extensive research~\cite{let,Talk} has focused on developing encoding strategies to render graph data compatible with LLMs. These strategies can primarily be classified into two branches: Graph2Text and Graph2Vec. The Graph2Text method translates graph data into textual sequences that preserve the structural and relational integrity in a sequential format suitable for direct ingestion by LLMs. Conversely, the Graph2Vec approach transforms graphs into vector representations, which are subsequently mapped into the embedding space of LLMs.

In this paper, we choose to utilize the Graph2Text method to convert graph data into a format suitable for processing by LLMs. Specifically, we describe the structure of the graph using triples of the form $(u, v, w)$, as elaborated in Section~\ref{sec:pre}. To illustrate, consider a graph from Figure~\ref{fig:overview} which can be represented as a sequence of such triples in~\ref{box:graph}:
\begin{TextBox}{Example: Graph Projection}\label{box:graph}
   This is a undirected graph, where $(u, v, w)$ denotes that node u and node v are connected by an undirected edge with the weight w. \\
(0, 2, 0.8) (0, 3, 0.4) (1, 2, 0.5) (2, 3, 0.3) 
\end{TextBox}
This structured textual representation preserve and convey the graph's relational information in a form that is comprehensible for LLMs, thereby facilitating efficient graph-based reasoning tasks. While alternative graph encoding approaches, such as using an adjacency matrix, are feasible and convenient. 



\paragraph{\textbf{Graph Reasoning Paths via Graph Algorithms}}

Accurate and diverse alignment data is crucial for graph reasoning tasks; however, collecting high-quality alignment data is non-trivial. Conventional methods, such as human or AI-powered annotation, exhibit distinct drawbacks. Manual annotation necessitates fundamental expertise in graph theory, leading to prohibitively high costs. Conversely, directly adopting AI tools for annotation fails to ensure accuracy. We analyze this phenomenon in our experiments in Section~\ref{sec:exp}, where even a powerful model like GPT-4 exhibits limited graph reasoning capability. To address this issue, we have devised a straightforward yet effective method to construct graph reasoning data. We utilize established graph algorithms to solve graph reasoning problems, meticulously recording both the solution processes and the answers. For example, we employ the Breadth-First Search (BFS) algorithm for connectivity tasks, Depth-First Search (DFS) for cycle detection tasks, and Dijkstra’s algorithm for shortest path tasks. By doing so, we guarantee the correctness of the answers. We refer to these directly obtained correct answers as Plain Answers (PA).

\begin{TextBox}{Example: PA}
[\textbf{Question}] Is there a path between node 0 and node 1? \\~
[\textbf{Plain Answer}] The answer is yes. 
\end{TextBox}

Recent work~\cite{expresssive} has shown that transformer decoders, without any intermediate steps, can only solve problems that lie within a fairly small circuit complexity class~\cite{merrill2023parallelism}. Therefore, constructing answers with intermediary steps is crucial. Existing studies~\cite{hsieh2023distilling,graphwiz} demonstrate that integrating CoT reasoning pathways can significantly enhance the reasoning capabilities of LLMs. One straightforward approach is to use a powerful LLM to generate reasoning paths when given a question and its corresponding Plain Answer (PA). However, as we will verify in Section~\ref{sec:exp}, even when provided with the correct answer, GPT-4 struggles to generate accurate reasoning paths. Fortunately, when utilizing graph algorithms to solve graph reasoning problems, we concurrently record both the answers and their solution processes. This allows us to leverage these solution processes to construct manually designed reasoning paths, which we denote as Answer with Hard Path (AHP).
\begin{TextBox}{Example: AHP}
[\textbf{Question}] Is there a path between node 0 and node 1? \\~
[\textbf{Answer with Hard Path}]  \textcolor{blue}{Node 0 is connected to node 2, node 2 is connected to node 1, we can follow the path: 0->2->1.} Yes, there a path between node 0 and node 1. 
\end{TextBox}


However, the manual construction of rule-based AHPs tends to result in uniform answer formats, potentially causing the model to overfit to this specific format~\cite{zhang2024enhancing}. This lack of diversity in training can lead to a decrease in the model's generalization performance. 
To address this issue, we employ LLMs to rewrite the AHP-generated answers, thereby enhancing their diversity. We refer to the answers obtained through this method as Answers with Soft Path (ASP).
It is noteworthy that when provided with AHP, the task facing the LLM shifts from graph reasoning to sentence rewriting. Sentence rewriting is inherently well-suited to LLMs due to their proficiency in language manipulation.
\begin{TextBox}{Example: ASP}
[\textbf{Question}] Is there a path between node 0 and node 1? \\~
[\textbf{Answer with Soft Path}] To find out if there is a path between node 0 and node 1, examine the connections given in the graph. \textcolor{blue}{Start from node 0, which directly connects to nodes 2 and 3. From node 2, the neighbors are 0, 3, and 1. Following this path, a connection exists from node 0 to node 2, and from node 2 to node 1.} Thus, the path from node 0 to node 1 is through node 2, confirming that a path exists between node 0 and node 1.
\end{TextBox}

This approach not only ensures the correctness of the reasoning process but also provides a structured method to enhance LLMs’ graph reasoning capability through accurate and logically sound CoT data.

\paragraph{\textbf{Alignment Tuning}}
We fine-tune LLMs using datasets that include graph reasoning paths to align them better with graph understanding and enable reasoning based on graph structures. Our training data set, $\mathcal{D}=\left \{ (\mathcal{G}_i,T_i, Q_i,R_i,A_i ) \right \} _{i=1}^N $, consists of $N$ quintuples where each element represents a graph $\mathcal{G}_i$, a task description $T_i$, a query $Q_i$, a reasoning path $R_i$, and an answer  $A_i$ respectively.

Initially, we employ the Graph Projection method to transform each graph $\mathcal{G}_i$ into a textual sequence, denoted as $ S_{\mathcal{G}_i} = f_P(\mathcal{G}_i)$ , where $f_P$ is the Graph Projection function. The reasoning path $R_i$ is obtained through graph algorithms, expressed as  $R_i = f_R(\mathcal{G}_i, T_i, Q_i)$, where $f_R$ is the function that uses graph algorithms to solve the graph reasoning task and yields the reasoning path.
The training objective is to maximize the probability of generating the correct answers based on this structured input:
\begin{equation}
\label{eq:max}
 \max_{G_\theta } p(A_i|S_{\mathcal{G}_i},T_i,Q_i,R_i ),   
\end{equation}
where $G_\theta $ is the parameter of \model. To enable \model ~to generate intermediate reasoning processes that aid in predicting the final answer, we formulate this as $(\hat{R_i}, \hat{A_i}) = f_G(S_{\mathcal{G}_i}, T_i, Q_i)$, where $f_G$ is the inference function of \model. The training objective function is defined as 
\begin{equation}
    \mathcal{L} = \mathcal{L}_A + \lambda \mathcal{L}_R,
\end{equation}
where $\mathcal{L}_{\text {A}}=\frac{1}{N} \sum_{i=1}^N \ell\left(\hat{A_i}, A_i\right)$ represents the answer prediction loss, and $\mathcal{L}_{\text {R}}=\frac{1}{N} \sum_{i=1}^N \ell\left(\hat{R_i}, R_i\right)$ is the reasoning path generation loss. $\ell$ is the cross-entropy loss between the predicted and target tokens and $\lambda$ is a hyperparameter.


\subsection{Theoretical Analysis}\label{sec:theory}

In this section, we provide a theoretical analysis of how CoT reasoning paths can enhance the reasoning capabilities of LLMs, thereby facilitating the generation of correct answers with greater ease. Due to the complexities of graph reasoning tasks, LLMs without robust reasoning abilities often fail to generate intermediate responses $Z$ for the ultimate correct answer $a$. As analyzed above, the utilization of an explicit CoT reasoning path $R$ can boost the LLM's reasoning ability, thus enabling more accurate outcomes.

\begin{theorem}
\label{theorem1}
Given the following conditions:
\begin{enumerate}
    \item  Non-triviality: The reasoning path $ R $ provides non-trivial information about the responses $ Z $, such that $ H(R|Z) > 0 $.
    \item Relevance: The reasoning path $ R $ contains information relevant to the correct answer $ a $ that is not fully captured by the response $ Z $, such that $I(a;R|Z)>0$.
\end{enumerate}
Then it follows that $ H(a|Z,R) < H(a|Z) $.    
\end{theorem}
In Theorem~\ref{theorem1}, $H(\cdot | \cdot)$ represent the conditional entropy and $I(\cdot;\cdot)$ denotes the mutual Information. The conditions coincides with observations from previous studies~\cite{expresssive,tighter} that point to LLMs' shortcomings in handling sequential reasoning challenges, such as simulating finite state machines, determining connectivity in graphs, or solving matrix equations. The reasoning  typically requires a series of logical steps and transformations that a simple direct model output $Z$ might not fully capture. We provide a proof of Theorem~\ref{theorem1} in Appendix~\ref{app:proof}.
In practice, the reasoning path $ R $ encapsulates progressive, structured reasoning or derivation steps leading to $ Z $. Therefore, knowing $ R $ reduces the uncertainty about $ a $ more effectively than knowing just $ Z $.
This theoretical analysis hinges on the nature of $ R $ providing supplementary, clarifying information beyond what $ Z $ alone offers, aligning with principles of information theory~\cite{information} where additional context reduces entropy.

\section{Experiment} \label{sec:exp}
In this section, we conduct experiments to address two critical research questions ($\mathcal{RQ}$):
\begin{itemize}
    \item $\mathcal{RQ}1$: How does \model~ perform in graph reasoning tasks compared to current SOTA open-source and closed-source LLMs? 
    \item $\mathcal{RQ}2$: What factors significantly impact the graph reasoning capabilities of LLMs?
\end{itemize}

\subsection{Dataset and Experimental Settings}~\label{se:exp_data}

\paragraph{\textbf{Dataset}} 


We conducted experimental validation on the NLGraph benchmark~\cite{wang2023can}, a comprehensive graph reasoning benchmark designed to evaluate performance across a spectrum of graph reasoning tasks. This benchmark encompasses eight distinct levels and complexities of tasks, namely: Connectivity, Cycle, Topological Sort, Shortest Path, Maximum Flow, Bipartite Graph Matching, Hamilton Path, and Graph Neural Networks. Detailed statistical information about the dataset and elaborate descriptions of the eight graph reasoning tasks are provided in Appendix~\ref{app:dataset}.

\paragraph{\textbf{Baselines}}

In our experimentation, we opted for a comparative analysis using both closed-source and open-source models as baselines. Specifically, we selected two closed-source models, GPT-4~\cite{gpt4} (version gpt-4-0125-preview) and GPT-3.5 (version gpt-3.5-turbo-1106), alongside two open-source models, Vicuna-7B~\cite{chiang2023vicuna} (version vicuna-7b-v1.5) and Llama3-8B~\cite{llama3} (version Llama 3.1 8B Instruct). Furthermore, we utilized Vicuna-7B and Llama3-8B as the base models for training two additional models, respectively denoted as \model-V and \model-L in our study.

\paragraph{\textbf{Settings}}

We follow the settings of NLGraph~\cite{wang2023can} for the dataset split, prompt, and the evaluation of the experimental results. Specifically, for all the eight tasks, we use accuracy as the evaluation metric. All experiments are conducted on an 8$*$A800 machine. The learning rate is set to 2e-5 and the hyperparameter $\lambda$ is set to 1.
More detailed experimental settings are available in Appendix~\ref{app:setting}.

\begin{table*}[htb]
\caption{Model performance under ZERO-SHOT.}
\label{tab:main_res}
\resizebox{\linewidth}{!}{
\begin{tabular}{@{}l|cccc|cccc|ccc|cccc@{}}
\toprule
          & \multicolumn{4}{c|}{Connectivity} & \multicolumn{4}{c|}{Cycle}           & \multicolumn{3}{c|}{Shortest Path}      & \multicolumn{4}{c}{Topology}  \\ \midrule
Model     & Easy   & Medium  & Hard   & Avg.  & Easy        & Medium & Hard  & Avg.  & Easy        & Hard        & Avg.        & Easy  & Medium & Hard & Avg.  \\ \midrule
GPT-3.5   & 73.21  & 69.39   & 55.46  & 65.50 & 80.00       & 55.14  & 59.32 & 59.69 & 34.48       & 8.57        & 20.31       & 79.31 & 20.51  & 0.00 & 28.89 \\
GPT-4     & 98.21  & 93.88   & 82.35  & 90.84 & {\ul 92.00} & 55.14  & 52.54 & 59.16 & {\ul 75.86} & {\ul 37.14} & {\ul 54.69} & 82.76 & 23.08  & 0.00 & 31.11 \\
Vicuna-7B & 67.86  & 59.69   & 51.26  & 58.22 & 52.00       & 49.53  & 59.32 & 52.88 & 6.90        & 0.00        & 3.12        & 0.00  & 0.00   & 0.00 & 0.00  \\
Llma3-8b  & 73.21  & 58.67   & 47.06  & 57.14 & 64.00       & 53.27  & 50.85 & 53.93 & 27.59       & 5.71        & 15.62       & 0.00  & 0.00   & 0.00 & 0.00  \\ \midrule
\model-V &
  \textbf{100.00} &
  {\ul 94.90} &
  {\ul 92.44} &
  {\ul 94.88} &
  {\ul 92.00} &
  {\ul 94.39} &
  {\ul 96.61} &
  {\ul 94.76} &
  68.97 &
  20.00 &
  42.19 &
  \textbf{96.55} &
  {\ul 42.31} &
  0.00 &
  {\ul 45.19} \\
\model-L &
  \textbf{100.00} &
  \textbf{100.00} &
  \textbf{98.32} &
  \textbf{99.46} &
  \textbf{96.00} &
  \textbf{96.26} &
  \textbf{98.31} &
  \textbf{96.86} &
  \textbf{82.76} &
  \textbf{42.86} &
  \textbf{60.94} &
  \textbf{96.55} &
  \textbf{70.51} &
  \textbf{10.71} &
  \textbf{63.70} \\ \bottomrule
\end{tabular}}
\resizebox{\linewidth}{!}{
\begin{tabular}{@{}l|ccc|ccc|ccc|ccc|c@{}}
\toprule
 &
  \multicolumn{3}{c|}{Flow} &
  \multicolumn{3}{c|}{Match} &
  \multicolumn{3}{c|}{Hamilton} &
  \multicolumn{3}{c|}{GNN} &
  \multirow{2}{*}{Overall Avg.} \\ \cmidrule(r){1-13}
Model     & Easy           & Hard       & Avg.  & Easy  & Hard  & Avg.  & Easy  & Hard & Avg.  & Easy  & Hard & Avg.  &       \\ \midrule
GPT-3.5   & 7.69           & 0.00       & 3.45  & 21.57 & 9.09  & 16.67 & 8.00  & 6.06 & 6.90  & 0.00  & 0.00 & 0.00  & 25.18 \\
GPT-4     & \textbf{23.08} & 0.00       & 10.34 & 58.82 & 15.15 & 41.67 & 20.00 & 3.03 & 10.34 & 56.52 & 0.00 & 33.33 & 41.43 \\
Vicuna-7B & 0.00           & 0.00       & 0.00  & 0.00  & 0.00  & 0.00  & 0.00  & 0.00 & 0.00  & 0.00  & 0.00 & 0.00  & 14.28 \\
Llama3-8b & 3.85           & {\ul 9.38} & 6.90  & 0.00  & 0.00  & 0.00  & 0.00  & 0.00 & 0.00  & 0.00  & 0.00 & 0.00  & 16.70 \\ \midrule
\model-V &
  \textbf{23.08} &
  6.25 &
  {\ul 13.79} &
  \textbf{66.67} &
  {\ul 21.21} &
  {\ul 48.81} &
  {\ul 40.00} &
  \textbf{27.27} &
  {\ul 32.76} &
  {\ul 82.61} &
  0.00 &
  {\ul 48.72} &
  {\ul 52.64} \\
\model-L &
  19.23 &
  \textbf{15.62} &
  \textbf{17.24} &
  {\ul 62.75} &
  \textbf{48.48} &
  \textbf{57.14} &
  \textbf{56.00} &
  \textbf{27.27} &
  \textbf{39.66} &
  \textbf{100.00} &
  \textbf{87.50} &
  \textbf{94.87} &
  \textbf{66.23} \\ \bottomrule
\end{tabular}}

\end{table*}

\subsection{Main Results $(\mathcal{RQ}1$)}\label{sec:exp_main}

We conducted experimental validations on eight graph reasoning tasks in a zero-shot setting. The accuracy for each difficulty level, as well as the average accuracy (Avg.), is provided in Table~\ref{tab:main_res}. Both open-source models, Vicuna-7B and Llama3-8B, exhibited poor performance across all tasks. Despite its advanced generative capabilities, the newly released Llama3-8B showed slightly better results than Vicuna-7B; however, both models demonstrated limited graph reasoning capabilities. The closed-source models, GPT-3.5 and GPT-4, displayed fundamental graph reasoning abilities. On simpler tasks such as Connectivity, their performance markedly surpassed that of the open-source models. However, when faced with complex graph reasoning tasks such as Maximum Flow and GNN, GPT-3.5 fell short. Although GPT-4 showed somewhat improved results over GPT-3.5, its performance was still suboptimal. \model-V and \model-L demonstrated commendable performance across all eight tasks, surpassing the advanced generative model GPT-4 with parameter sizes of 7B and 8B, respectively. They exceeded their corresponding base models by 38.36\% and 49.53\%, validating the effectiveness of our proposed reasoning path construction and Alignment Tuning in enhancing LLMs' capabilities in graph reasoning. This indicates that LLMs are inherently capable of graph reasoning but require appropriate alignment with graph reasoning tasks to exhibit this capacity. Notably, \model-L achieved exceptionally high accuracy in the GNN task, suggesting that LLMs can effectively simulate two-layer graph convolution operations on relatively small-scale graphs, which involve complex multi-step reasoning.

        
        
        

\begin{figure*}[htp]
  \centering
  \subfigure[Connectivity.]{
		\label{fig:Connectivity}
		\includegraphics[width=0.32\textwidth]{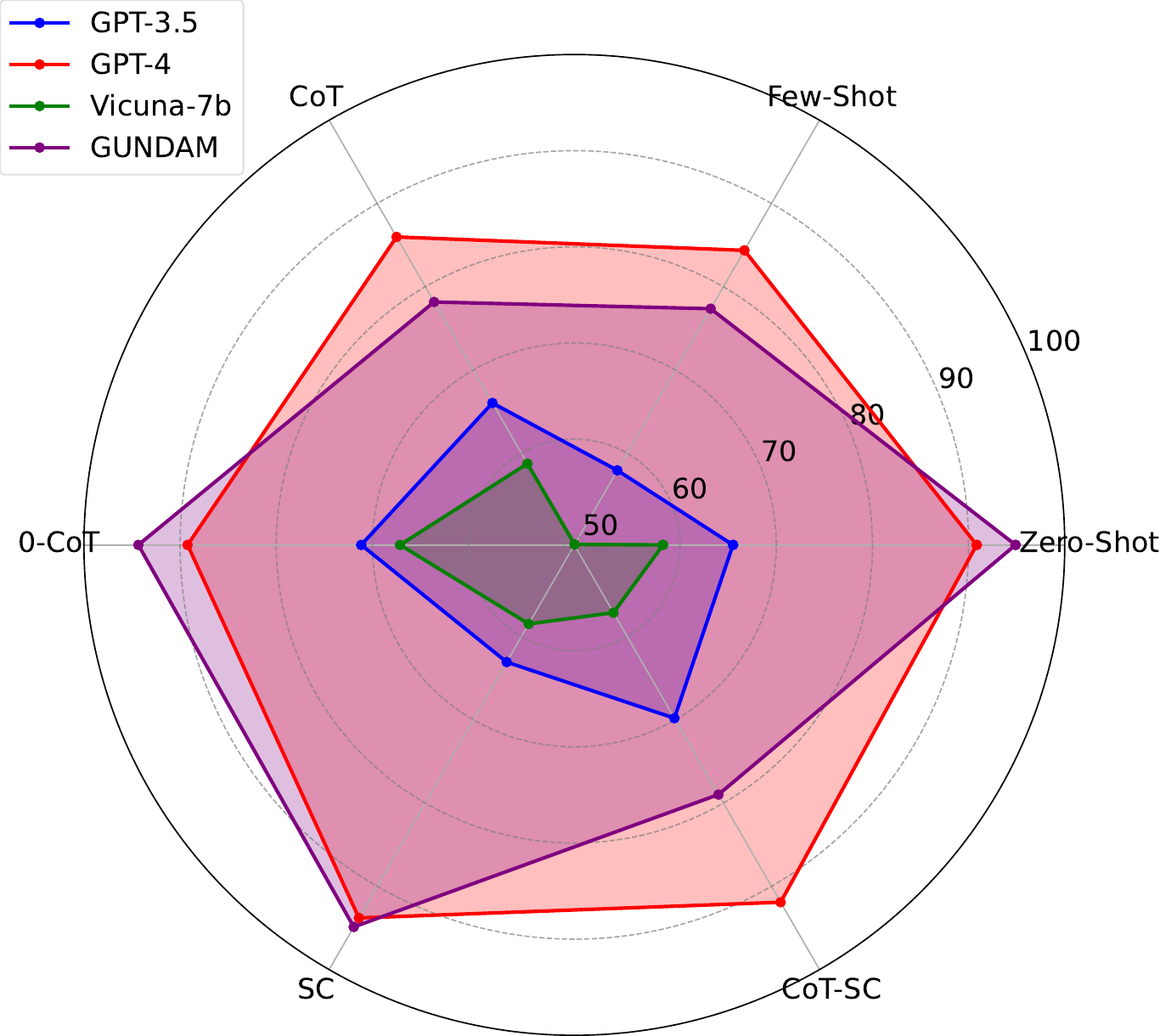}}
    \subfigure[Cycle.]{
		\label{fig:Cycle}
		\includegraphics[width=0.32\textwidth]{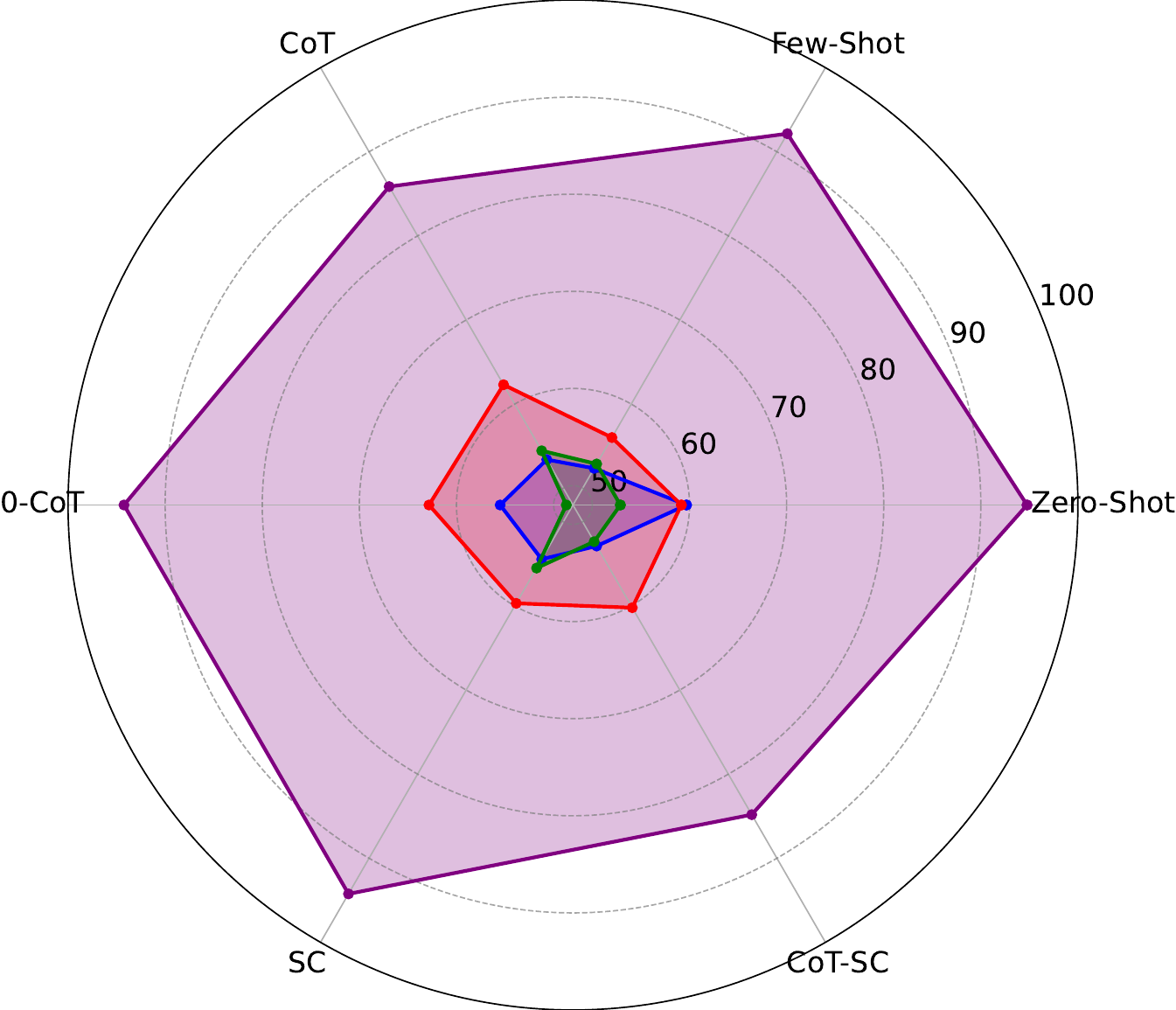}}		
    \subfigure[Shortest Path.]{
		\label{fig:Shortest}
		\includegraphics[width=0.32\textwidth]{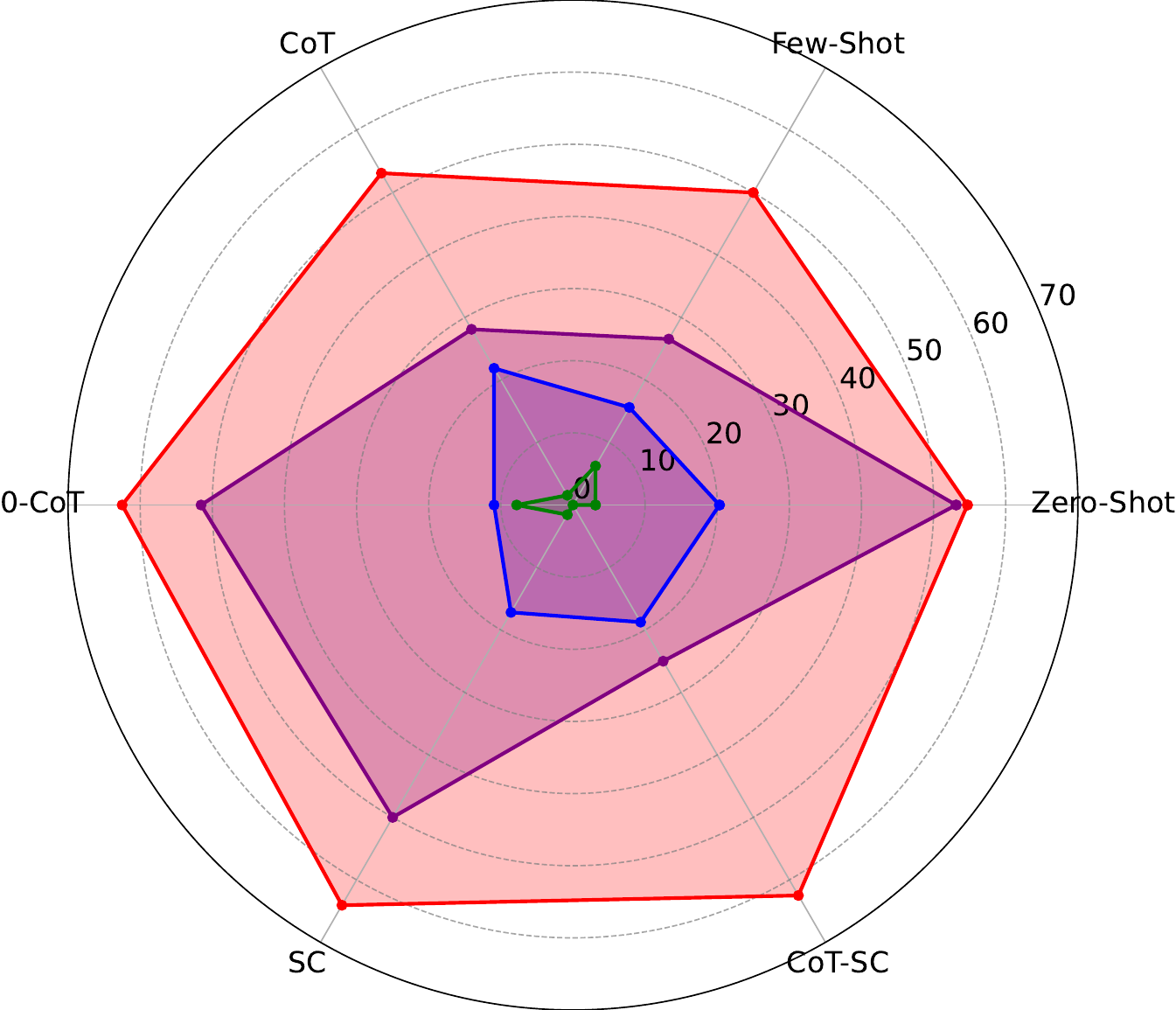}}		
    
    \caption{Comparison of the performance for different prompt techniques.} 
    \label{fig:prompt}
\end{figure*}

\subsection{Ablation Study ($\mathcal{RQ}2$)}~\label{se:exp_ablation}

In this section, we aim to investigate the impact of reasoning paths, prompt techniques, and training data difficulty on the reasoning abilities. Unless specified otherwise, all subsequent experiments use Vicuna-7B as the base model, i.e., \model-V.

\paragraph{\textbf{Reasoning Path}}
\begin{table}[htb]
\centering
\caption{Impact of reasoning path.}
\label{tab:ab_path}
\begin{tabular}{@{}l|ccc@{}}
\toprule
       & Connectivity & Cycle & Shortest Path \\ \midrule
Random & 50.00        & 50.00 & 5.54          \\ \midrule
PA     & 55.53        & 55.50 & 25.00         \\
AGP    & 87.33        & 56.02 & 12.50         \\
AHP    & 91.11        & 93.72 & 37.50         \\
ASP    & 91.11        & 96.61 & 39.06         \\ \bottomrule
\end{tabular}
\end{table}

In this subsection, we further investigate how reasoning paths influence the graph reasoning performance of LLMs. We conducted experiments on the Connectivity, Cycle, and Shortest Path tasks, with the average accuracy presented in Table~\ref{tab:ab_path}. More detailed results are provided in Appendix~\ref{app:path}. We use AGP to denote Answers with Generated Path, where reasoning paths are generated by GPT-4. Additionally, following NLGraph~\cite{wang2023can}, we include a Random baseline, which randomly assigns ``Yes'' or ``No'' to the Connectivity and Cycle tasks, with an expected accuracy of 50\%. For the Shortest Path task, we randomly pick a valid path and the sum of the weights along the path as the answer.

The results from models trained on PA, AGP, Answers with Hard Path (AHP), and ASP surpassed random outcomes, indicating that aligning LLMs with data containing correct answers generally enhances their reasoning capabilities to various extents. PA showed moderate performance across the three tasks, suggesting that data with only answers and no intermediate processes provide limited improvement in LLMs' reasoning abilities. AGP performed well in the Connectivity task but was less effective in the Cycle and Shortest Path tasks. This is likely because AGP's reasoning paths, generated by GPT-4, are reliable in simpler tasks such as Connectivity, where GPT-4 can generate correct paths. However, GPT-4 struggles with more complex reasoning tasks, leading to incorrect reasoning paths and, consequently, incorrect answers in the Cycle and Shortest Path tasks.
Both AHP and ASP consistently demonstrated superior performance across all tasks, reaffirming that integrating reasoning paths can indeed strengthen LLMs' reasoning abilities, provided these paths are accurate. This also confirms that our methodology for constructing reasoning paths is effective. Furthermore, while AHP and ASP achieved comparable performance, models aligned with ASP data could generate more diverse responses, showcasing the added benefit of ASP in enhancing response variability.

\paragraph{\textbf{Prompt}}


We further explored the impact of prompt techniques on the graph reasoning performance of LLM. We conducted experiments using five distinct prompt techniques, with results depicted in Figure~\ref{fig:prompt}. Typically, zero-shot configurations yielded relatively good or even the best results. However, advanced prompting techniques such as CoT and SC resulted in negligible performance improvements. These findings suggest that merely employing prompt techniques struggles to significantly enhance LLMs' understanding and reasoning capabilities on graphs. In contrast, our proposed Alignment Tuning offers a viable solution to effectively bolster the graph reasoning abilities of LLMs. This insight highlights the limitations of conventional prompting strategies and underscores the importance of tailor-made adaptations such as Alignment Tuning to fully leverage LLMs in complex reasoning tasks involving graph-structured data.

\paragraph{\textbf{Data Difficulty}}

\begin{figure}[H]
  \centering
  \subfigure[Easy to Hard.]{
		\label{method_subgraph_instance}
		\includegraphics[width=0.4\textwidth]{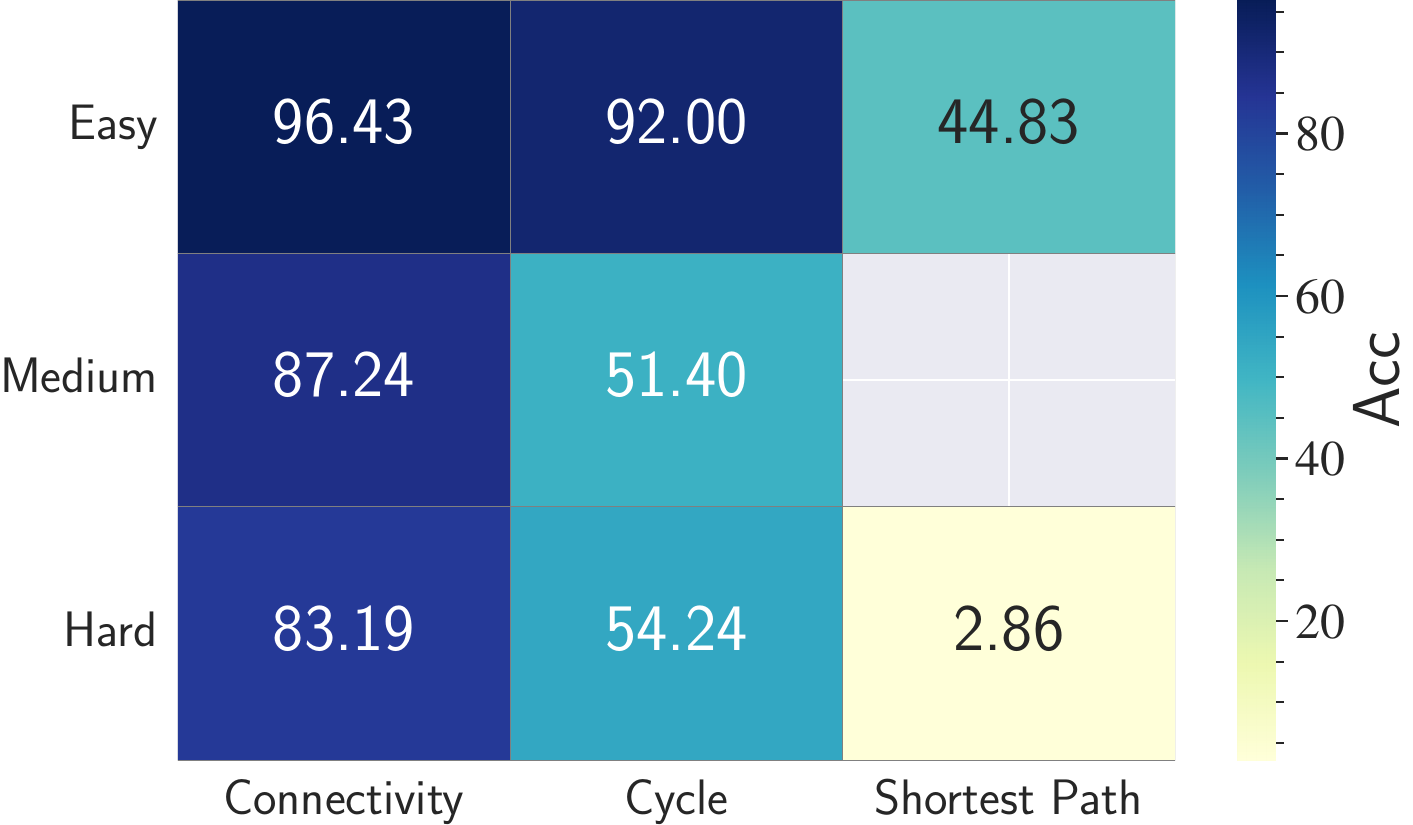}}
    \subfigure[Hard to Easy.]{
		\label{method_node_edge_graph}
		\includegraphics[width=0.4\textwidth]{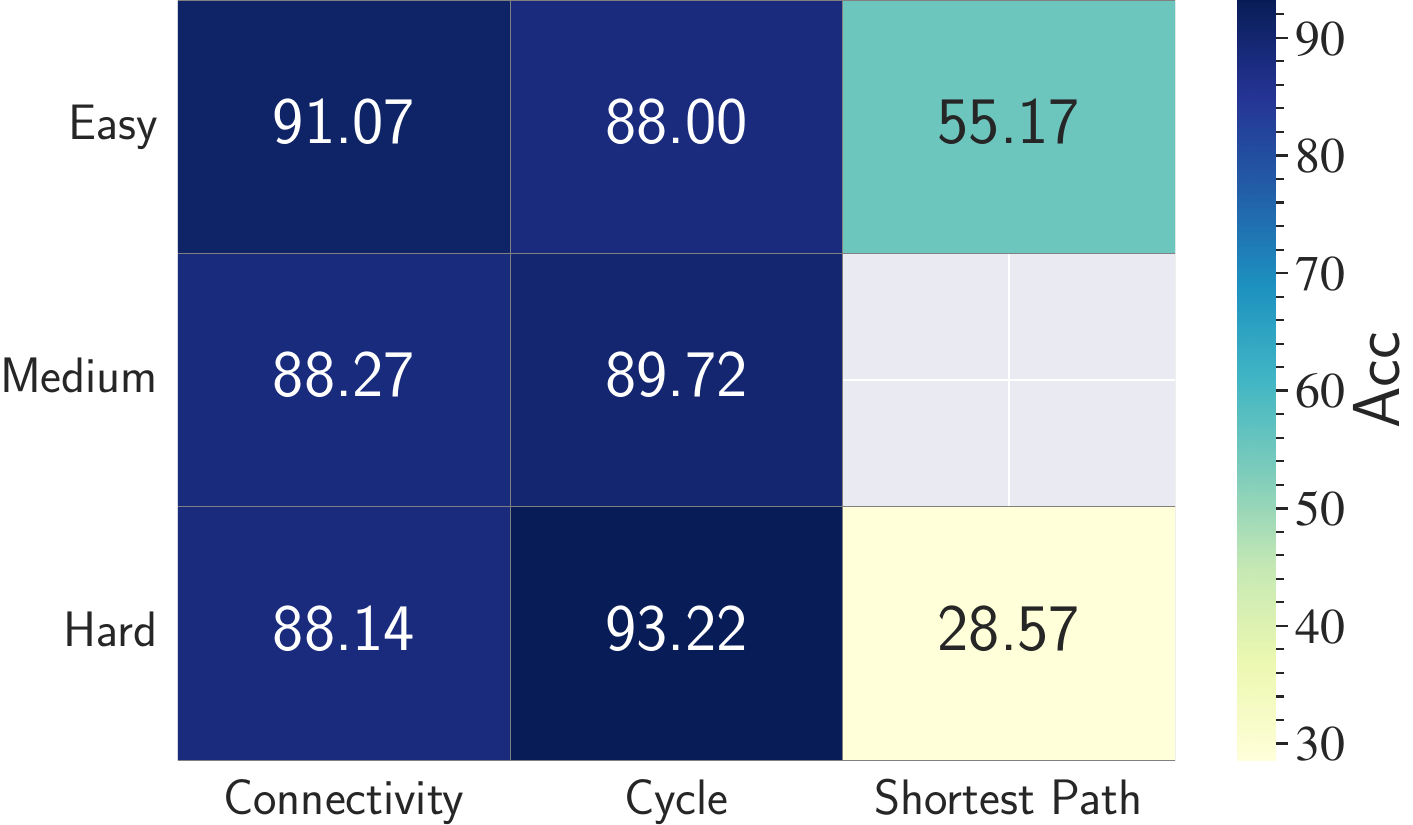}}
    \caption{Transfer Experiment.} 
    \label{fig:transfer}
  \setlength{\abovecaptionskip}{0cm}
\end{figure}

\begin{figure*}[htb]
    \centering
    \includegraphics[width=0.8\linewidth]{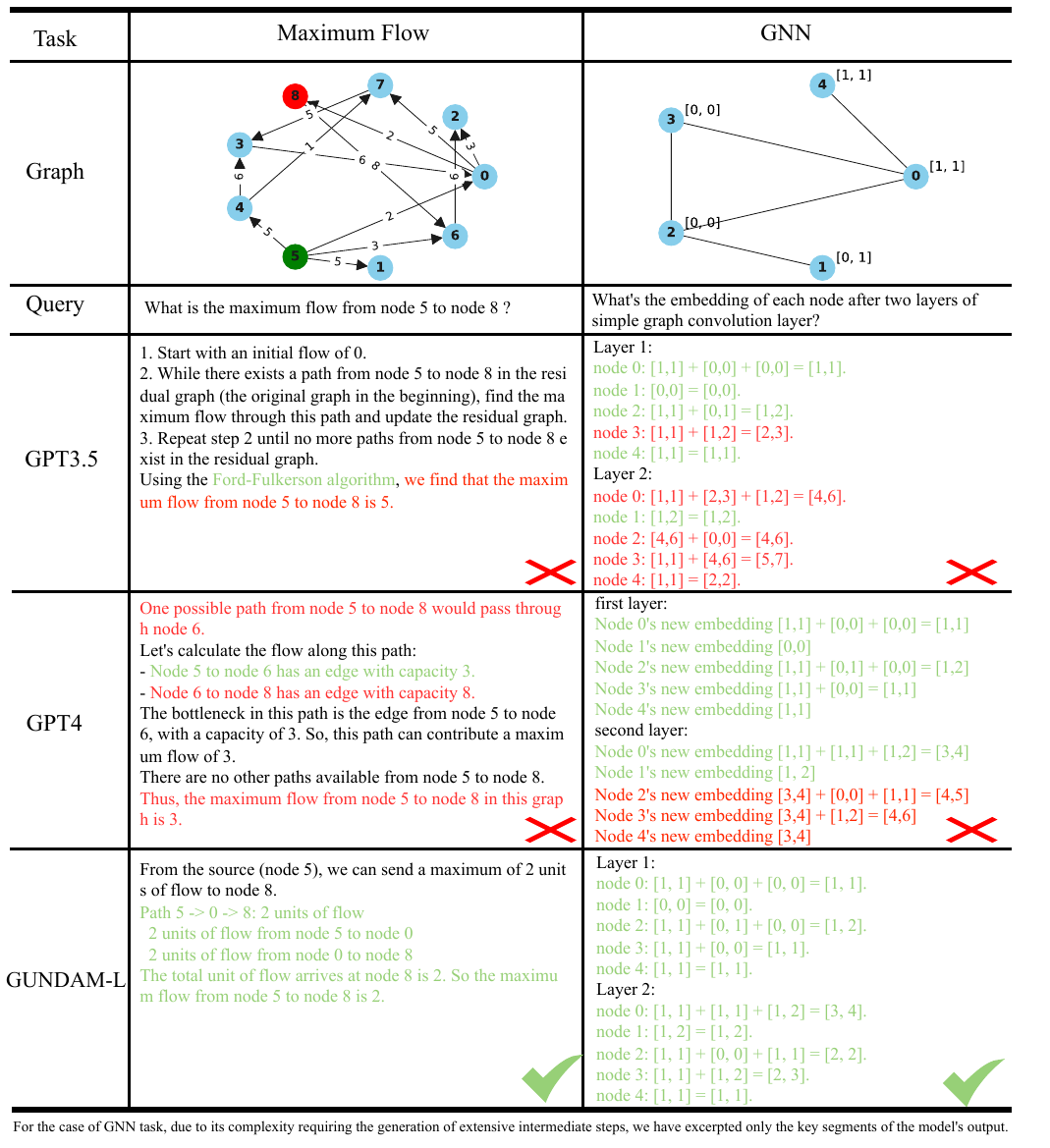}
    \vspace{-0.2cm}
    \caption{Case study.}
    \label{fig:case}
    \vspace{-0.2cm}
\end{figure*}

To investigate the impact of data difficulty on model performance, we conducted transfer experiments using datasets of varying difficulties. Specifically, we implemented two scenarios: 1) easy to hard, where we trained models with easy-level data and tested them on datasets of increasing difficulty to evaluate the transferability of reasoning capabilities; and 2) hard to easy, where we trained models on hard-level data and tested them on on varying difficulties. The results are presented in Figure~\ref{fig:transfer}.

For simpler reasoning tasks such as Connectivity, models trained on easy data demonstrated effective transferability to other difficulty levels. However, for more complex reasoning tasks, the effectiveness of transfer from easy data was limited. Conversely, models trained on hard data showed significant transferability across all tested difficulty levels, suggesting that using hard data in training substantially enhances the LLM's graph reasoning capabilities. This effectiveness may be attributed to the intricate reasoning processes invoked by hard data, which likely induces the model to generate more intermediate steps, thereby enhancing reasoning depth and robustness. This observation aligns with findings in \citet{expresssive}, which analyzed how the length of a transformer’s chain of thought impacts its reasoning power.
Moreover, while models trained solely on hard data performed better than those trained on easy data, they underperformed compared to models trained on a mixed-difficulty dataset. This indicates that, although hard data contributes significantly to enhancing reasoning capabilities, a hybrid training approach incorporating datasets of various difficulties yields the best results. This mixture presumably builds a more robust and versatile reasoning model, capable of handling both simple and complex graph reasoning tasks effectively.

\subsection{Case Study}~\label{se:exp_case}

In Figure~\ref{fig:case}, we present case studies for two complex tasks: Maximum Flow and GNN. Results from the two base models are omitted for these tasks due to their inability to handle such complexity.
For the Maximum Flow task, GPT-3.5 identified an algorithm capable of solving the problem but failed to detail the necessary intermediate processes, resulting in an incorrect answer. GPT-4, on the other hand, offered more detailed intermediate results but erroneously reversed the direction of the edge from node 8 to node 6, leading to incorrect intermediate steps and consequently a wrong conclusion. Conversely, \model-L not only provided correct intermediate steps but also accurately reasoned the correct answer.

Regarding the GNN task, which requires modeling two layers of graph convolution operations involving complex multi-step reasoning, GPT-3.5 made errors in simulating even the first layer of convolution. It incorrectly updated the embedding of node 3 using the post-update embedding of node 2 as if they were the original embedding. GPT-4 accurately modeled the first layer but replicated the same error as GPT-3.5 in the second layer, wrongly using the updated embedding of node 0 for updating nodes 2, 3, and 4 in that layer, leading to incorrect final embedding for these nodes. \model-L, however, executed correct feature updates at each layer, resulting in the correct outcomes.
From these case studies, it is apparent that GPT-4 exhibits stronger reasoning capabilities than GPT-3.5 but still encounters challenges with tasks requiring complex and multi-step reasoning, occasionally leading to erroneous reasoning outcomes. In contrast, \model-L demonstrated robust graph reasoning proficiency, effectively understanding and manipulating complex graph structures through multi-step reasoning. Furthermore, \model-L not only produces correct answers but also generates a reasoning path that elucidates how those answers were derived, enhancing the interpretability of the model’s outputs.


\section{Conclusion}
To enhance the ability of LLMs to understand graph data and perform reasoning tasks based on graph structures, we introduce \model. It employs a Graph Projection method to convert graph structures into textual formats that LLMs can process and constructs CoT reasoning data via graph algorithms. Furthermore, we propose Alignment Tuning to effectively align LLMs with graph reasoning tasks. Experiments conducted on eight graph reasoning tasks demonstrate the efficacy of \model, validating its utility in enabling sophisticated graph-based reasoning capabilities in LLMs.

\bibliographystyle{unsrtnat}
\bibliography{references}  

\appendix
\onecolumn
\section{Proof of theorem} \label{app:proof}
\begin{theorem}
Given the following conditions:
\begin{enumerate}
    \item  Non-triviality: The reasoning path $ R $ provides non-trivial information about the responses $ Z $, such that $ H(R|Z) > 0 $.
    \item Relevance: The reasoning path $ R $ contains information relevant to the correct answer $ a $ that is not fully captured by the response $ Z $, such that $I(a;R|Z)>0$.
\end{enumerate}
Then it follows that $ H(a|Z,R) < H(a|Z) $.    
\end{theorem}

\paragraph{Definitions and Notation}
\begin{itemize}
    \item $H(\cdot|\cdot) $ denotes the conditional entropy.
    \item $I(\cdot;\cdot)$ denotes the mutual Information.
    \item $ Z $ denotes the responses generated by the LLM.
    \item $ R $ denotes the reasoning path.
    \item $ a $ denotes the correct answer.
\end{itemize}


\paragraph{Intuition}
The main idea is that the reasoning path $ R $ provides additional information about the correct answer $ a $ beyond what is provided by the response $ Z $. Therefore, having access to $ R $ should reduce the uncertainty about $ a $, hence reducing the conditional entropy.

\begin{proof}

Consider the chain rule of conditional entropy:

\[
H(a, R | Z) = H(a | Z) + H(R | a, Z)
\]
and
\[
H(a, R | Z) = H(R | Z) + H(a | R, Z).
\]

Setting these two expressions equal, we obtain:

\[
H(a | Z) + H(R | a, Z) = H(R | Z) + H(a | R, Z).
\]




\[
H(a | Z) -H(a | R, Z)  = H(R | Z) -H(R | a, Z)= I(a;R|Z).
\]

Given $ I(a;R|Z) \ge 0 $:

\[
H(a | Z) -H(a | R, Z) = I(a;R|Z) \ge 0.
\]

We have proven that:

\[
H(a | Z) > H(a | R, Z),
\]



Thus, this completes the proof that having access to the reasoning path $ R $ in addition to the responses $ Z $ reduces the uncertainty about the correct answer $ a $, proving $ H(a|Z,R) < H(a|Z) $.
\end{proof}

\section{Details of training}

Constructing Answers with Soft Paths (ASP) essentially involves a sentence rewriting task, an area in which LLMs demonstrate substantial proficiency. For this purpose, researchers can utilize any commonly available open-source or proprietary LLM. In this paper, we specifically employ GPT-4 to undertake this task, owing to its advanced text generation and transformation capabilities. To facilitate ASP construction using GPT-4, we employ two commonly used prompts. These prompts strategically guide GPT-4 to reformulate the initial Answer with Hard Path (AHP) into a more varied and generalized form, enhancing the diversity and robustness of the model outputs. 
\begin{TextBox}{Prompt Templates for ASP}
\textbf{Template1.} \\~
Given an question [question] and its answer [answer], derive the shortest yet complete Chain of Thought reasoning path that leads to the answer. \\~
[begin of question] \\ ~
\{question\} \\ ~
[end of question] \\~
[begin of answer] \\~
\{answer\} \\~
[end of answer] \\~
 \hrule 
 ~\\~
\textbf{Template2.} \\~
Rewrite the following sentence to improve its logical flow: \\~
\{answer\}
\end{TextBox}



\section{Dataset} \label{app:dataset}
\subsection{Dataset statistic.}
\begin{table*}[htb]
\caption{Dataset statistic.}
\label{tab:data}
\begin{threeparttable}
\begin{tabular}{@{}ccccccccc@{}}
\toprule
Subset   & Connectivity & Cycle    & Topo. Sort & Shortest Path & Max. Flow & Matching & Hamilton & GNN     \\ \midrule
\#Easy   & 296/56       & 125/25   & 151/29     & 151/29        & 124/26    & 249/51   & 125/25   & 77/23   \\
Size     & n: 5-10      & n: 5-10  & n: 5-10    & n: 5-10       & n: 5-10   & n: 6-20  & n: 5-10  & n: 5-8  \\ \midrule
\#Medium & 1004/196     & 493/107  & 372/78     & /             & /         & /        & /        & /       \\
Size     & n: 11-25     & n: 11-25 & n: 11-25   & /             & /         & /        & /        & /       \\ \midrule
\#Hard   & 561/119      & 341/59   & 152/28     & 165/35        & 168/32    & 177/33   & 167/33   & 124/16  \\
Size     & n: 26-35     & n: 26-35 & n: 26-35   & n: 11-20      & n: 11-20  & n: 17-33 & n: 11-20 & n: 9-15 \\ \midrule
Total    & 1861/371     & 959/191  & 675/135    & 316/64        & 292/58    & 426/84   & 292/58   & 201/39  \\ \bottomrule
\end{tabular}
 \begin{tablenotes}
        \footnotesize
        \item 
         Size denotes the number of nodes in graphs. A/B denotes the number of samples in the training/test set respectively.
      \end{tablenotes}
      \vspace{-0.1cm}
  \end{threeparttable}
\end{table*}
\subsection{Reasoning Task} \label{app:task}
\begin{itemize}
    \item \textbf{Connectivity}: Given an undirected graph $\mathcal{G}$ and two nodes $u$ and $v$, determine whether there is a path to connect them, following $u\rightarrow v_i \rightarrow \cdots  \rightarrow v_j \rightarrow v$.
    \item \textbf{Cycle}: Given an undirected graph $\mathcal{G}$, determine whether there is a cycle in the graph, following $u\rightarrow v_i \rightarrow \cdots  \rightarrow v_j \rightarrow u$.
    \item \textbf{Shortest Path}: Given a weighted undirected graph $\mathcal{G}$ and two nodes $u$ and $v$, the shortest path is the one that connects the two nodes and minimizes the sum of the weights of the edges, such that $$P = \arg\min_{P \in \mathcal{P}{uv}} \sum_{(v_i, v_j) \in P} w_{ij},$$
    where $\mathcal{P}_{uv}$ denotes the set of all possible paths from $u$ to $v$ in the graph $\mathcal{G}$, and $(v_i, v_j) \in P$ represents each edge in the path $P$ with weight $w_{ij}$.
    \item \textbf{Topological Sort}: A topological sort of a directed graph arranges its nodes in a sequence such that for every directed edge $(u,v)$ from node $u$ to node $v$, node $u$ precedes node $v$ in the sequence. The goal is to determine a valid topological order for a given directed graph, acknowledging that there may be multiple correct solutions. We task LLMs with producing a valid topological sort for the given graph and then use an external program to verify its accuracy.
    \item \textbf{Maximum Flow}: Consider a network $\mathcal{G}=\{\mathcal{V},\mathcal{E}\}$ with two specific nodes $s$ and $t$ representing the source and sink, respectively. Each edge in the network has an associated capacity $c$, which indicates the maximum flow that can pass through it. The task is to instruct LLMs to devise a strategy to maximize the flow from the source to the sink. The evaluation criteria include both exact matches with the optimal solution and partial credit. The partial credit is calculated as follows: 
    \[
\mathrm{PC} = \begin{cases} 
\frac{t}{s}, & \text{if } t \leq s \\
0, & \text{if } t > s 
\end{cases},
\]
where $s$ represents the flow value in the optimal plan, and $t$ is the flow value in the solution generated by the LLMs.
    \item \textbf{Bipartite Graph Matching}: In an undirected graph $\mathcal{G}=\{\mathcal{V},\mathcal{E}\}$, a matching refers to a collection of edges such that no two edges share a common node. A bipartite graph is a special type of graph where the set of nodes can be partitioned into two disjoint sets $\mathbf{U}$ and $\mathbf{V}$, with no edges between nodes within the same set. The goal is to find a matching in the bipartite graph that includes the maximum possible number of edges. We then use an external program to assess whether the solution generated by the LLMs is both valid and optimal.
    \item \textbf{Hamilton Path}: A Hamilton path in an undirected graph is a route that passes through each node exactly one time. For a given undirected graph $\mathcal{G}=\{\mathcal{V},\mathcal{E}\}$, the objective is to identify a valid Hamilton path. We preprocess the base graphs to guarantee the existence of at least one Hamilton path and then use an external program to assess the solution provided by the LLM.
    \item \textbf{Graph Neural Networks}: Given an undirected graph $\mathcal{G}=\{\mathcal{V},\mathcal{E}\}$ and a two-dimensional embedding $\mathbf{x}_i$ for each node, the goal is to execute $\ell$ layers of message passing. This involves updating each node's embedding by summing the embeddings of all its neighboring nodes. Formally, this update is represented as:
\[
\mathbf{x}_{i}^{(\ell+1)} =  \sum_{j \in {\mathcal{N}_{i}}} \mathbf{x}_{j}^{(\ell)},
\]
where $\mathcal{N}_{i}$ represents the set of neighbors for node $i$, and $(\ell)$ indicates the $\ell$-th layer. The evaluation for this task includes exact matches with the correct node embeddings and two types of partial credits. The first partial credit is based on the percentage of nodes with correct embeddings (PC), while the second involves the average of the relative errors across all embedding dimensions (RE). The relative error is calculated as $RE = \frac{|x-y|}{\max(x,y)}$, where $x$ is the value generated by the LLMs, and $y$ is the corresponding value in the standard answer, averaged over all embedding dimensions.
\end{itemize}

\section{Experiment Setting}\label{app:setting}

For the FEW-SHOT based prompt techniques (including FEW-SHOT, CoT and CoT-SC), the input prompt contains $K$ examples of pertinent questions and answers. For Connectivity and Cycle tasks, $K$ is 4 and $K$ is set to 5 for Shortest Path task. For the self-consistency (SC) prompt method, we sample 5 responses. All experiments are conducted on an 8*A800 machine. We employ Vicuna-7B and Llama3-8B as the base model for \model, the learning rate is set to 2e-5, and the batchsize is set to 8. For evaluation, the temperature factor $\tau$ is set to 0.2 except for the self-consistency Prompt method, whose $\tau$ is 0.8.

\section{Supplementary Experiment} \label{app:exp}

\subsection{Reasoning Path}\label{app:path}
\begin{table}[H]
\caption{ Impact of reasoning path.}
\label{tab:app_path}
\begin{tabular}{@{}l|cccc|cccc|ccc@{}}
\toprule
       & \multicolumn{4}{c|}{Connectivity} & \multicolumn{4}{c|}{Cycle}    & \multicolumn{3}{c}{Shortest Path} \\ \midrule
       & Easy   & Medium  & Hard   & Avg.  & Easy & Medium & Hard  & Avg.  & Easy      & Hard      & Avg.      \\ \midrule
random & 50     & 50      & 50     & 50    & 50   & 50     & 50    & 50    & 6.07      & 5.08      & 5.54      \\ \midrule
PA     & 53.57  & 51.53   & 63.03  & 55.53 & 56   & 54.21  & 57.63 & 55.5  & 41.38     & 11.43     & 25        \\
AGP    & 92.86  & 87.76   & 84.03  & 87.33 & 56   & 59.81  & 49.15 & 56.02 & 20.69     & 5.71      & 12.5      \\
AHP    & 100    & 91.84   & 85.71  & 91.11 & 96   & 92.52  & 94.92 & 93.72 & 62.07     & 17.14     & 37.5      \\
ASP    & 96.43  & 90.82   & 89.08  & 91.11 & 96   & 93.46  & 96.61 & 94.76 & 51.72     & 28.57     & 39.06     \\ \bottomrule
\end{tabular}
\end{table}

\subsection{Case Study}\label{app:case}
We provide further insights with two case studies illustrated in Table~\ref{tab:case_connectivity}. 
For the first case, considering the graph on the left. All four models correctly indicated the presence of a path, assuring the connectivity between the two nodes. However, the Vicuna-7B made an error in its reasoning process by suggesting a non-existent connection from node 17 to node 16. GPT-3.5, GPT-4, and \model~ provided correct answers with different valid paths. Although these paths were not the shortest, they were indeed existent and verified the connectivity as required.
In the second case, both GPT-4 and \model~ accurately concluded the absence of any connecting path, whereas GPT-3.5 and Vicuna-7B generated incorrect responses. GPT-3.5 mistakenly introduced a non-existent edge (6,4), and Vicuna-7B incorrectly proposed an edge (6,5). In analyzing other error cases of GPT3.5, it was also found that GPT3.5 tends to construct non-existent edges to make two nodes connected when no path exists in the graph. 
\begin{table*}[htb]
\caption{Cases from Connectivity Task.}
\label{tab:case_connectivity}

\begin{threeparttable}
\resizebox{\linewidth}{!}{
\begin{tabular}{@{}c|l|l@{}}
\toprule
Graph &
  \multicolumn{1}{c|}{\raisebox{-.5\height}{\includegraphics[width=0.2\linewidth]{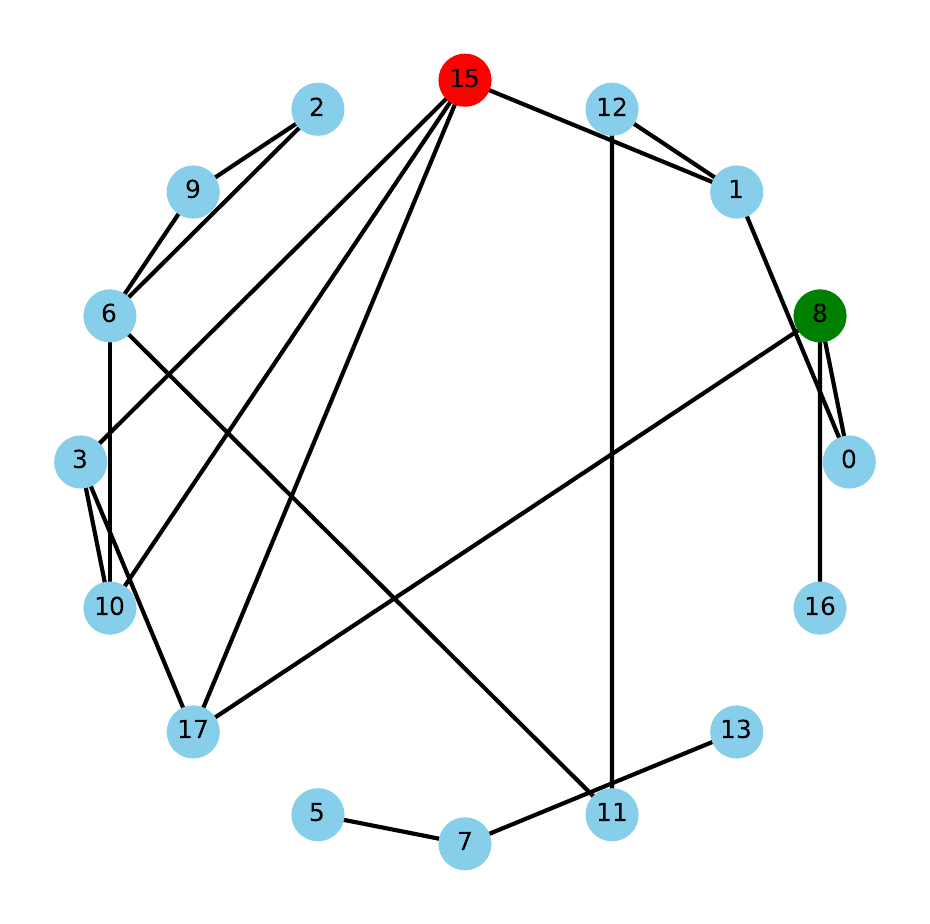}}} &
  \multicolumn{1}{c}{\raisebox{-.5\height}{\includegraphics[width=0.2\linewidth]{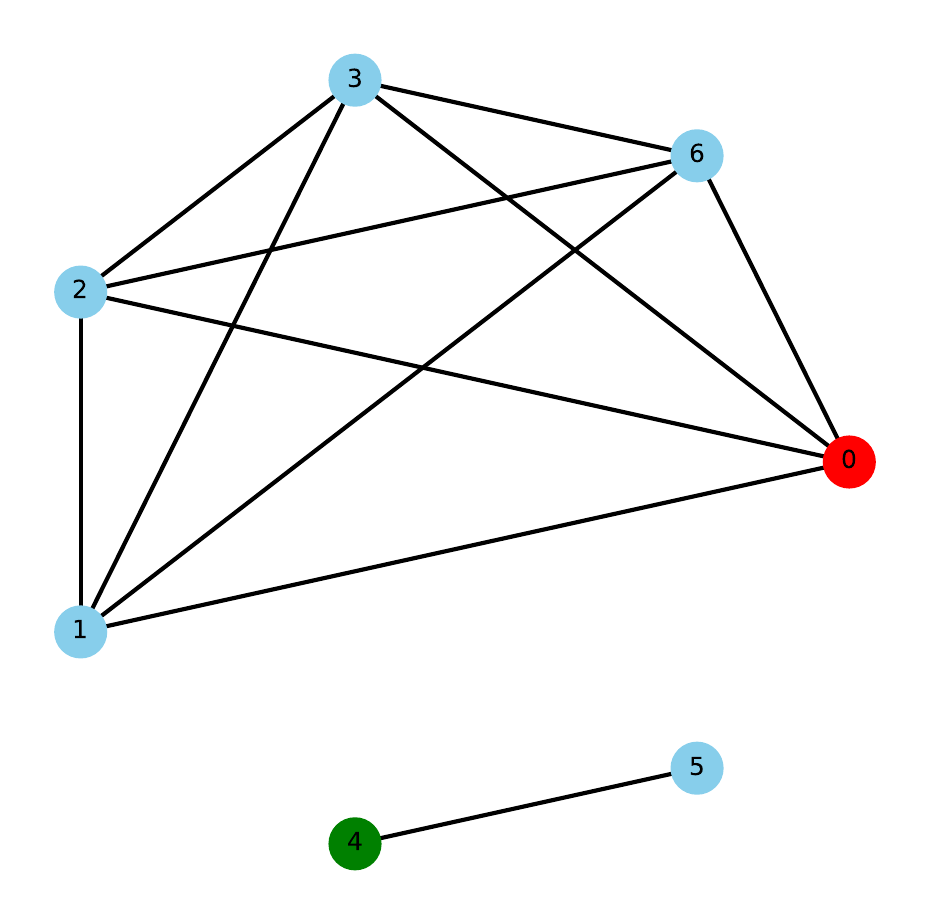}}} \\ \midrule
Query &
  Is there a path between node 15 and node 8? &
  Is there a path between node 0 and node 4? \\ \midrule
GPT3.5 &
  \begin{tabular}[c]{@{}l@{}}Yes, there is a path between node 15 and node 8\\ in the provided graph. The path is as follows: 15-1-0-8\end{tabular} &
  \begin{tabular}[c]{@{}l@{}}\textcolor{red}{Yes, there is a path between node 0 and node 4 in the given}\\ \textcolor{red}{graph.} We can see that the following path connects these\\ two nodes: \textcolor{red}{0-3-6-1-2-3-6-4}.\end{tabular} \\ \midrule
GPT4 &
  \begin{tabular}[c]{@{}l@{}}Yes, there is a path between node 15 and node 8. \\ The path can be as follows: 15-3-17-8.\end{tabular} &
  \begin{tabular}[c]{@{}l@{}}No, there is no path between node 0 and node 4 in the given\\ graph. The nodes 0 and 4 are in separate disconnected subgraphs.\end{tabular} \\ \midrule
Vicuna-7B &
  \begin{tabular}[c]{@{}l@{}} \textbackslash{}***\textbackslash{\tnote{1}}~{ } Starting from node 15, we can follow the  connections\\to node 17, \textcolor{red}{then to node 16, and finally to node 8.}\\ Therefore, there is a path between node 15 and node 8.  \end{tabular} &
  \begin{tabular}[c]{@{}l@{}}\textbackslash{}***\textbackslash{\tnote{1}}~{ }  Starting from node 0, we can follow the edges as follows:\\ \textcolor{red}{(0,6) (6,3) (3,2) (2,1) (1,6) (6,5) (5,4)} \textcolor{red}{Therefore, the answer is Yes,}\\ \textcolor{red}{there is a path between node 0 and node 4.}\end{tabular} \\ \midrule
\model-V &
  \begin{tabular}[c]{@{}l@{}}Node 15 is connected to node 1, node 1 is connected to\\ node 0, node 0 is connected to node 8,  we can follow the path:\\ 15-1-0-8. Yes, there a path between node 15 and node 8.\end{tabular} &
  No, there is no path between node 0 and node 4. \\ \bottomrule
\end{tabular}}
 \begin{tablenotes}
        \footnotesize
        \item[1] \textbackslash{}***\textbackslash ~indicates that unimportant generated content has been omitted.
      \end{tablenotes}
      
  \end{threeparttable}
  
\end{table*}

\end{document}